\documentclass[10pt,twocolumn,letterpaper]{article}
\usepackage{wacv}
\usepackage{times}
\usepackage{epsfig}
\usepackage{graphicx}
\usepackage[T1]{fontenc}
\usepackage{savesym}
\savesymbol{iint}
\savesymbol{iint}
\usepackage{txfonts}
\restoresymbol{TXF}{iint}
\restoresymbol{TXF} {iint}

\newcommand{\minus}{\scalebox{0.75}[1.0]{$-$}}

\usepackage{amssymb}

\def\wacvPaperID{574}
\wacvfinalcopy
\ifwacvfinal
\def\assignedStartPage{1} 
\fi
\ifwacvfinal
\usepackage[breaklinks=true,bookmarks=false]{hyperref}
\else
\usepackage[pagebackref=true,breaklinks=true,colorlinks,bookmarks=false]{hyperref}
\fi

\ifwacvfinal
\else
\fi

\begin{document}

\title{LVRNet: Lightweight Image Restoration for \\ 
Aerial Images under Low Visibility}

\author{Esha Pahwa\thanks{equal contribution}\\
BITS Pilani\\
{\tt\small f20180675@pilani.bits-pilani.ac.in}
\and
Achleshwar Luthra\footnotemark[1]\\
Carnegie Mellon University\\
{\tt\small achleshl@andrew.cmu.edu}
\and
Pratik Narang\\
BITS Pilani\\
{\tt\small pratik.narang@pilani.bits-pilani.ac.in}
}
\maketitle
\begin{figure*}[ht]
    \centering
    \includegraphics[width = \textwidth]{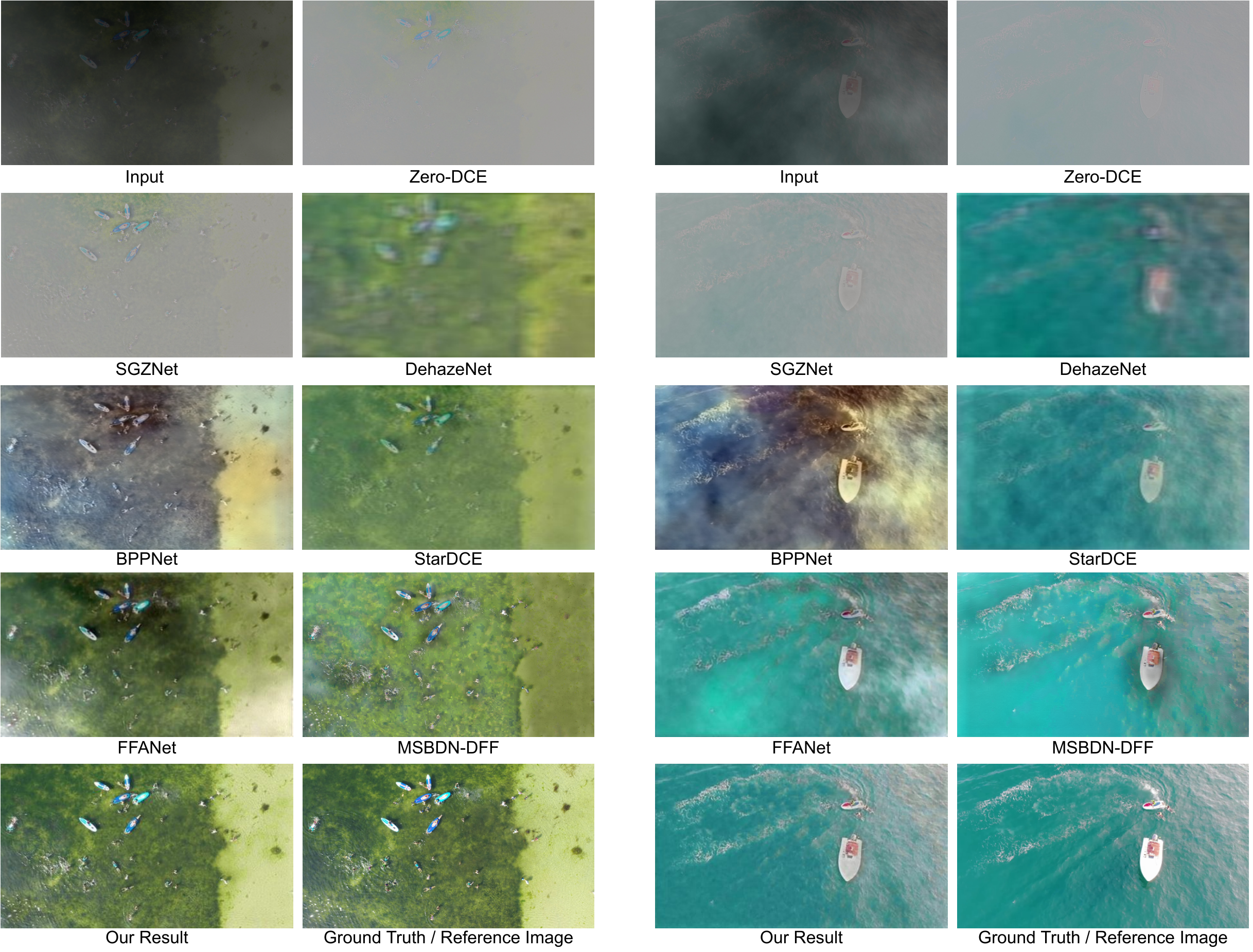}
    \caption{\textbf{Visual results on the proposed LowVis-AFO dataset.} The method used to obtain each result has been mentioned under the image.}
    \label{fig:abstract}
\end{figure*}

\begin{abstract}
\noindent Learning to recover clear images from images having a combination of degrading factors is a challenging task. That being said, autonomous surveillance in low visibility conditions caused by high pollution/smoke, poor air quality index, low light, atmospheric scattering, and haze during a blizzard becomes even more important to prevent accidents. It is thus crucial to form a solution that can result in a high-quality image and is efficient enough to be deployed for everyday use. However, the lack of proper datasets available to tackle this task limits the performance of the previous methods proposed. To this end, we generate the LowVis-AFO dataset, containing 3647 paired dark-hazy and clear images. We also introduce a lightweight deep learning model called Low-Visibility Restoration Network (LVRNet). It outperforms previous image restoration methods with low latency, achieving a PSNR value of 25.744 and an SSIM of 0.905, making our approach scalable and ready for practical use. The code and data can be found \href{https://github.com/Achleshwar/LVRNet}{here}.

\end{abstract}

\section{Introduction}

\noindent Image enhancement and restoration have been a critical area of research using both traditional digital image processing techniques\cite{geman1984stochastic} \cite{besag1991bayesian}, and the recent deep learning frameworks\cite{nah2017deep}\cite{qin2020ffa}\cite{zhang2017beyond}. The goal of image restoration is to recover a clear image, whereas  image enhancement is to improve the quality of the degraded image. In this study, we perform recovery of the clear image from the hazy version while performing low-light image enhancement using a single convolutional network, which could further be applied to tasks such as search and rescue operations using object detection.

\noindent Using deep learning algorithms for image recovery has many benefits, the most important being that it can generalize to different variations in the images captured. Hence, we observe that deep learning-based methods on most benchmark datasets often outperform traditional methods significantly. However, there are still challenges that the researchers have to tackle for image restoration. Publicly available datasets containing a variety of degrading factors that model real-world scenarios are few. Hence, most previous works have focused on removing one type of degradation with a specific intensity level. From the perspective of computational complexity, recent deep learning methods are computationally expensive, and thus they can't be deployed on edge devices. Moreover, image restoration has been a long-standing ill-posed research problem, as there are infinite mappings between the degraded and the clear image. Thus, the existing methods still have room for improvement in finding the correct mapping.

\noindent 

\noindent In this work, we focus on developing an end-to-end lightweight deep-learning solution for the image restoration task. Our major contributions are listed below:
\begin{itemize}
    \item Taking inspiration from Non-linear Activation Free Network (NAFNet) \cite{chen2022simple}  and Level Attention Module \cite{zhang2021benchmarking}, we propose a novel algorithm - Low-Visibility Restoration Network (LVRNet), that can effectively recover high-quality images from degraded images taken in poor visual conditions (\figurename~\ref{fig:abstract}). 
    \item Due to the lack of available datasets that exhibit a combination of adverse effects, we generate a new dataset, namely LowVis-AFO (abbreviation for Low-Visibility Aerial Floating Objects dataset). We use AFO \cite{afo} as our ground truth dataset and synthesize dark hazy images. The data generation process has been elaborated in Section~\ref{dataset}.
    \item Benchmarking experiments have been provided on the LowVis-AFO dataset to help future researchers for quantitative comparison. Along with that, LVRNet surpasses the results obtained using previous image restoration techniques by a significant margin. 
    \item  We perform extensive  ablation studies to analyze the importance of various loss functions existing in current image restoration research. These experiments are discussed in detail in Section ~\ref{res}.
\end{itemize}

\begin{figure*}[ht]
    \centering
    \includegraphics[width=0.8\textwidth, height=200pt]{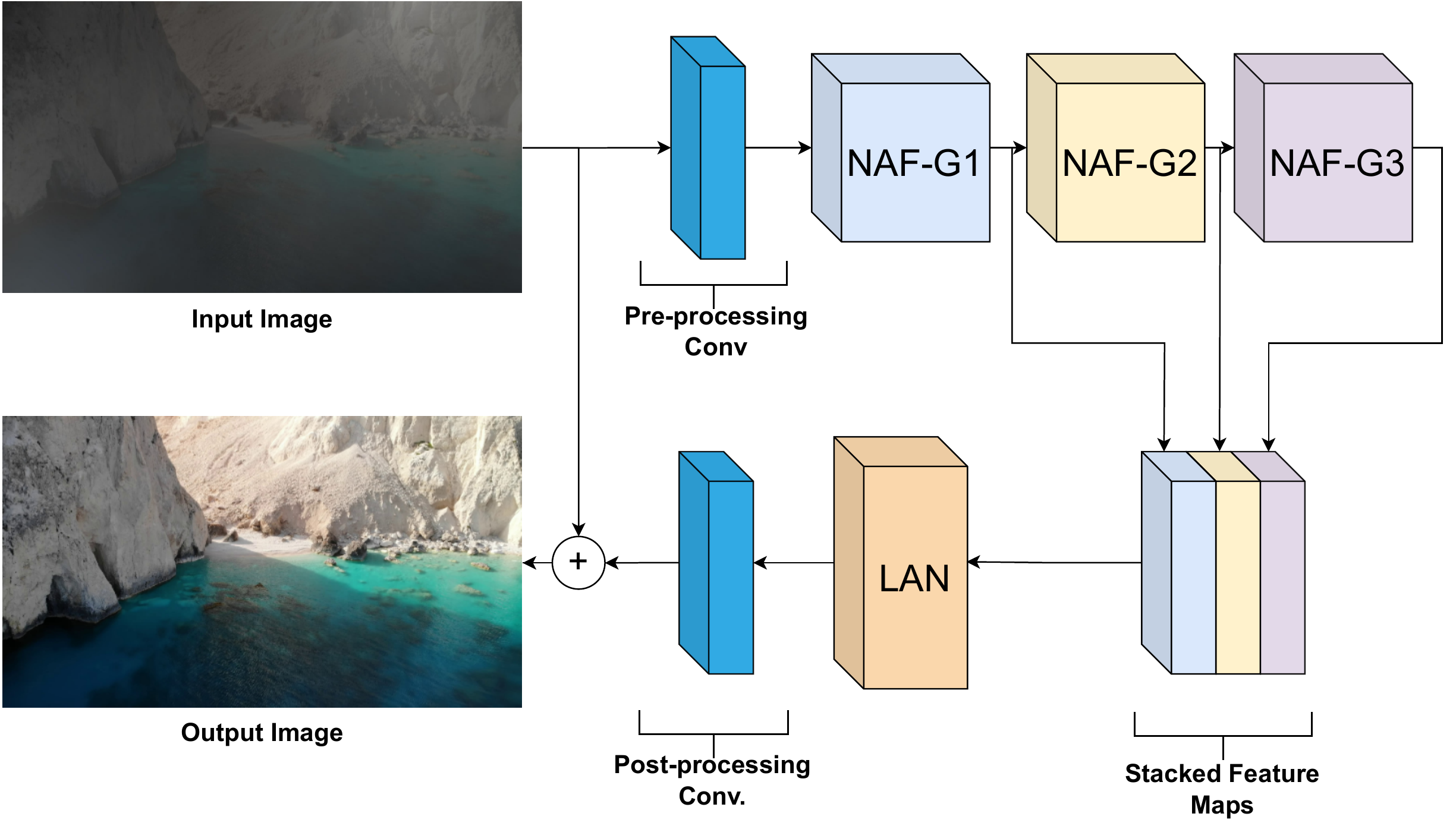}
    \caption{\textbf{Model architecture of the proposed LVRNet.} Starting from the top-left: The input image is passed to the pre-processing convolution layers where feature maps are learned and passed to NAF Groups (here we have used 3 groups). The features extracted from each group are concatenated (or stacked) along the channel dimension and sent as input to the Level Attention Module (LAM). Finally, we pass LAM's output to CNN layers for post-processing, adding the original image through residual connection and extracting the restored image at the bottom-left.}
    \label{fig:arch}
\end{figure*}

\section{Related Works}
\label{relatedwork}
\noindent This section highlights the previous work done in the fields of image dehazing and low-light image enhancement and their limitations. 

\subsection{Image Dehazing}
\noindent Hazy weather is often seen due to floating particles in the environment which degrade the quality of the image captured. Therefore, many previous works have tried to recover a clear image from the hazy one. These works can be divided into two methods, ones that rely on prior assumptions \cite{he2010single} and the atmospheric scattering model (ASM) \cite{mccartney1976optics} and the others which use deep learning to solve the problem, either by combination with ASM \cite{cai2016dehazenet}\cite{ren2016single}\cite{ren2020single} or independently \cite{liang2019selective}\cite{liu2019griddehazenet}\cite{qin2020ffa}\cite{zhang2021hierarchical}\cite{zheng2021ultra}.   
\noindent Conventional approaches are physically inspired and apply various types of sharp image priors to regularize the solution space. However, they exhibit shortcomings when implemented with real-world images and videos. For example, the dark channel prior method (DCP) \cite{mccartney1976optics} does not perform well in regions containing the sky. These methods \cite{ancuti2010fast}\cite{fattal2014dehazing}\cite{li2015simultaneous} are known to be computationally expensive and require heuristic parameter-tuning. Supervised dehazing methods can be divided into two subparts, one is ASM based, and the other is non-ASM based.
\\

\noindent \textbf{ASM-based Learning:} MSCNN\cite{ren2016single} solves the task of image dehazing by dividing the problem into three steps: using CNN to estimate the transmission map $t(x)$, using statistical methods to find atmospheric light $A$ and then recover the clear image $J(x)$ using $t(x)$ and $A$ jointly. Methods like LAP-Net \cite{li2019lap} adopt the relation of depth with the amount of haze in the image. The farther the scene from the camera, the denser the haze would be. Hence it considers the difference in the haze density in the input image using a stage-wise loss, where each stage predicts the transmission map from mild to severe haze scenes. DehazeNet \cite{cai2016dehazenet} consists of four sequential operations: feature extraction, multi-scale mapping, calculating local extremum, and non-linear regression. 
MSRL-DehazeNet \cite{yeh2019multi} decomposes the problem into recovering high-frequency and basic components. GCANet \cite{chen2019gated} employs residual learning between haze-free and hazy images as an optimization objective. 
\\

\noindent \textbf{End-to-end Learning:} This subpart of previous work corresponds to non-ASM-based deep learning methods for recovering the clear image. Back-Projected Pyramid Network (BPPNet) \cite{https://doi.org/10.48550/arxiv.2008.06713} is a generative adversarial network that includes iterative blocks of UNets \cite{https://doi.org/10.48550/arxiv.1505.04597} to learn haze features and pyramid convolution to preserve spatial features of different scales. The reason behind using iterative blocks of UNets\cite{https://doi.org/10.48550/arxiv.1505.04597} is to avoid increasing the number of encoder layers in a single UNet\cite{https://doi.org/10.48550/arxiv.1505.04597} as it leads to a decrease in height and width of latent feature representation hence resulting in loss of spatial information. Moreover, different blocks of UNet learn different complexities of haze features, and the final concatenation step ensures that all of them are taken into account during image reconstruction. The final reconstruction is done using the pyramid convolution block. The output feature is post-processed to get a haze-free image. Feature-Fusion Attention Network (FFANet) \cite{qin2020ffa} adopts the idea of an attention mechanism and skip connections to restore haze-free images. A combination of channel attention and pixel attention is introduced, which helps the network, deal with the uneven spatial distribution of haze and different weighted information across channels. Autoencoders \cite{chen2019convolutional}, hierarchical networks \cite{das2020fast}, and dense block networks \cite{guo2019dense123} has also been proposed for the task of image dehazing. However, our main comparison lies with FFANet \cite{qin2020ffa}, wherein we show a huge improvement compared to the former method with a model containing a lesser number of parameters and which can generalize to different levels of haze.

\begin{figure*}[ht]
    \centering
    \includegraphics[width=0.85\textwidth, height=150pt]{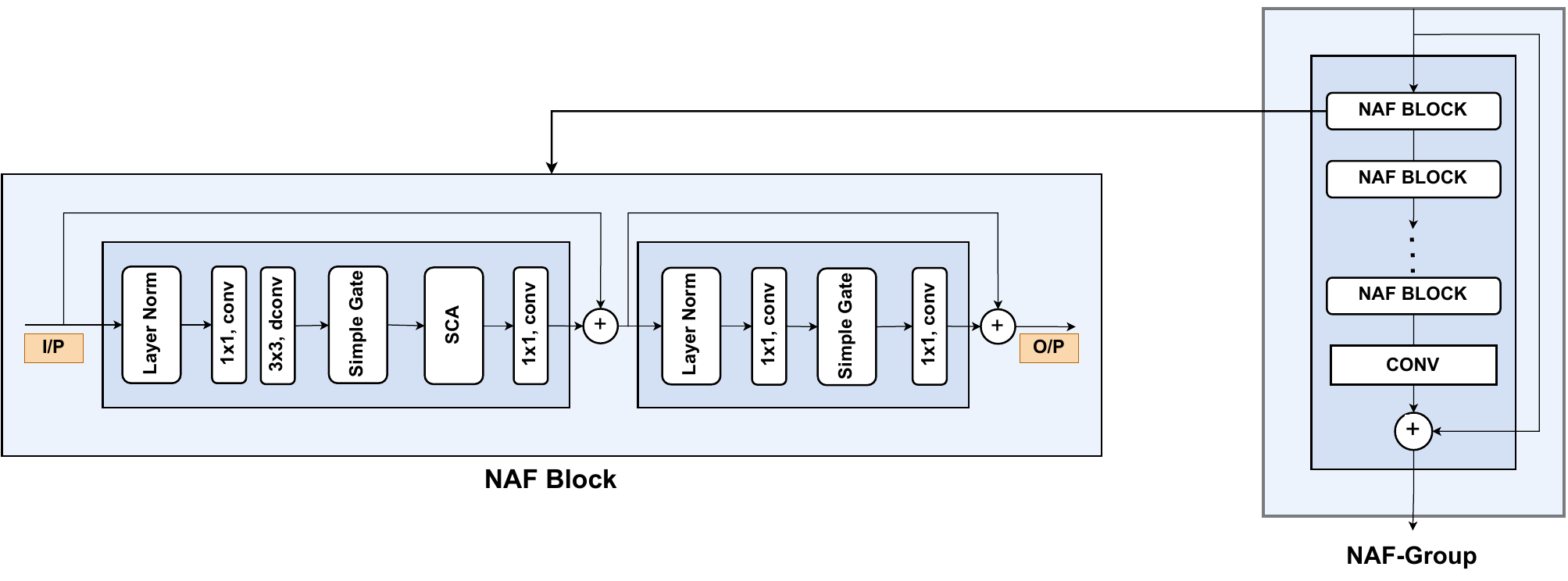}
    \caption{\textbf{Architecture of NAF Block and NAF Group.} NAF Blocks are the building blocks of NAF Groups. A detailed description has been provided in Section \ref{arch} and Section \ref{naf-block}}
    \label{fig:naf}
\end{figure*}
\subsection{Low-light Enhancement}
\noindent Traditional methods for low-light image enhancement (LLIE) include Histogram Equalization-based methods and Retinex model-based methods. Recent research has been focused on developing deep learning-based methods following the success of the first seminal work. Deep learning-based solutions are more accurate, robust, and have a shorter inference time thus attracting more researchers. Learning strategies used in these methods are mainly supervised learning \cite{lore2017llnet, lv2020fast, lv2018mbllen, ren2019low, zhu2020eemefn, lu2020tbefn, xu2020learning}, unsupervised learning \cite{jiang2021enlightengan}, and zero-shot learning \cite{zheng2022semantic, guo2020zero}.
\\
\noindent \textbf{Supervised Learning:} The first deep learning-based LLIE method LLNet \cite{lore2017llnet} is an end-to-end network that employs a variant of stacked-sparse denoising autoencoder to brighten and denoise low-light images simultaneously. LLNet inspired many other works \cite{lv2020fast, lv2018mbllen, ren2019low, zhu2020eemefn}, but they do not consider the observation that noise exhibits different levels of contrast in different frequency layers. Later, Xu et al. \cite{xu2020learning} proposed a network that suppresses noise in the low-frequency layers and recovers the image contents by inferring the details in high-frequency layers. There is another division of methods that is based on the Retinex theory. Deep Retinex-based models \cite{wei2018deep, yang2021sparse} decomposes the image into two separate components - light-independent reflectance and structure-aware smooth illumination. The final estimated reflection component is treated as the enhanced result. 
\\
\noindent\textbf{Unsupervised Learning:} Although the above-mentioned methods perform well on synthetic data, they show limited generalization capability on real-world low-light images. This might be the result of overfitting. EnlightenGAN \cite{jiang2021enlightengan} proposed to solve this issue by adopting an unsupervised learning technique, i.e., avoiding the use of paired synthetic data. This work uses attention-guided UNet as a generator and global-local discriminators to achieve the objective of LLIE. 
\\
\noindent \textbf{Zero-short Learning:} These methods, in low-level vision tasks, do not require any paired or unpaired training data. Zero-reference Deep Curve Estimation \cite{guo2020zero} formulates image enhancement as a task of image-specific deep curve estimation, taking into account pixel value range, monotonicity, and differentiability. It is a lightweight DCE-Net that doesn’t require paired or unpaired ground truth images during training and relies on non-reference loss functions that measure the enhancement quality hence driving the learning of the network. Another such method, Semantic-guided Zero-shot low-light enhancement Network \cite{zheng2022semantic} is a lightweight model for low-light enhancement factor extraction which is inspired by the architecture of U-Net \cite{https://doi.org/10.48550/arxiv.1505.04597}. The output of this network is fed to a recurrent image enhancement network, along with the degraded input image. Each stage in this network considers the enhancement factor and the output from the previous scale as its input. This is followed by a feature-pyramid network that aims to preserve the semantic information in the image.

\noindent More recently, researchers have experimented with transformers for Zero-shot Learning LLIE. Structure-Aware lightweight Transformer (STAR) \cite{zhang2021star} focuses on real-time image enhancement without using deep-stacked CNNs or large transformer models. STAR is formulated to capture long-range dependencies between separate image patches, facilitating the model to learn structural relationships between different regions of the images. In STAR, patches of the image are tokenized into token embeddings. The tokens generated as an intermediate stage are passed to a long-short-range transformer that outputs two long and short-range structural maps. These structural maps can further predict curve estimation or transformation for image enhancement tasks. Although these methods show impressive results for the study of low-light image enhancement for which it originally developed, they cannot deal with foggy low-- light images.

\subsection{Limitations}
\noindent Previous works have relied on ASM-based methods in the case of dehazing and Retinex model-based methods for low-light image enhancement. However, these methods fail to generalize to real-world images. Recent deep learning-based methods using large networks solve the task of image dehazing and low-light enhancement separately. To our knowledge, no work is introduced that solves the two problems in a collaborative network. Deep learning methods also fail to generalize to different haze levels and darkness.

\section{Proposed Methodology}

\noindent In this section, we provide a detailed description of the overall architecture proposed and the individual components included in the network.
\subsection{Architecture}\label{arch}

\noindent Like the group structure in \cite{qin2020ffa}, each group in our network consists of a $K$ NAF Block \cite{chen2022simple} with a skip connection at the end as shown in \figurename~\ref{fig:naf}. The output of each group is concatenated, passed to the level attention module to find the weighted importance of the feature maps obtained, and post-processed using two convolutional layers. A long skip connection for global residual learning accompanies this.

\subsubsection{NAF-Block} \label{naf-block}
To keep this work self-contained, we explain the NAF Block \cite{chen2022simple} in this subsection. NAF Block is the building block of Nonlinear Activation Free Network. Namely NAFNet \cite{chen2022simple}. To avoid over-complexity in the architecture, this block avoids using any activation functions like ReLU, GELU, Softmax, etc.\ hence keeping a check on the intra-block complexity of the network. 

\noindent The input first passes through Layer Normalization as it can help stabilize the training process. This is followed by convolution operations and a Simple Gate (SG). SG is a variant of Gated Linear Units (GLU) \cite{dauphin2017language} as evident from the following equations~\ref{glu} and \ref{sg}

\begin{equation}
    \label{glu}
    GLU(X, f, g, \sigma) = f(X) \odot \sigma(g(X))
\end{equation}

\begin{equation}
    \label{sg}
    SimpleGate(X, Y) = X \odot Y
\end{equation}

\noindent and a replacement for GELU\cite{hendrycks2016gaussian} activation function because of the similarity between GLU and GELU (Equation~\ref{gelu}).

\begin{equation}
    \label{gelu}
    GELU(x) = x \phi (x)
\end{equation}

\begin{figure}
    \centering
    \includegraphics[width=0.4\textwidth]{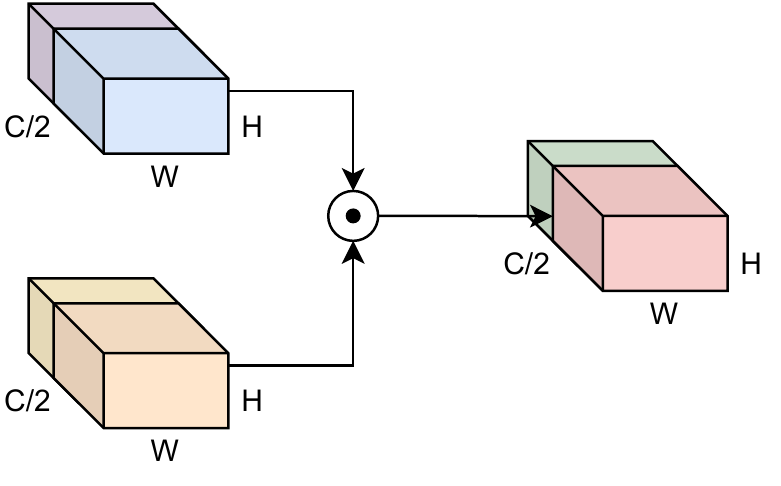}
    \caption{Simple Gate as represented by Equation \ref{sg} $\otimes$ denotes channel-wise multiplicaWere}
    \label{fig:sg}
\end{figure}

\noindent In Simple Gate, the feature maps are divided into two parts along the channel dimension and then multiplied as shown in \figurename~\ref{fig:sg}. Another novelty introduced in this block is Simplified Channel Attention (SCA). Channel Attention (CA) can be expressed as: 

\begin{equation}
    \label{ca}
    CA(X) = X \otimes \sigma(W_2 max(0, W_1 pool(X)))
\end{equation}

\noindent where $X$ represents the feature map, $pool$ indicates the global average pooling operation,$\sigma$ is Sigmoid, $W1, W2$ are fully-connected layers and $\otimes$ is a channel-wise product operation. This can be taken as a special case of GLU from which we can derivate the equation for Simplified Channel Attention: 

\begin{equation}
    \label{sca}
    SCA(X) = X \otimes W pool(X)
\end{equation}

\subsubsection{Level Attention Module}
Once we have extracted features from all the NAF Groups, we concatenate them and pass them through the Level Attention Module (LAM) \cite{zhang2021benchmarking}. This module learns attention weights for features obtained at different levels. 

\noindent In LAM, each feature map is first reshaped to a 2D matrix of the size $K \times HWC$, where $K, H, W,$ and $C$ are the no. of NAF Groups, height, width, and no. of channels of the feature maps respectively. We find a correlation matrix of this 2D matrix by multiplying it with its transpose matrix. Finally, we multiply the 2D matrix with this correlation matrix and reshape it to $K \times H \times W \times C$ tensor. Inspired by residual learning, this tensor is substituted for residual and is added to the original concatenated feature maps. The resultant features are then reshaped to $H \times W \times KC$, passing through $1 \times 1$ convolution operation to get the $H \times W \times C$ feature map. This is passed through some post-processing convolutions to get the final enhanced output. We include its architecture diagram in the supplementary material for a better understanding.

\subsection{Loss Functions}
\label{lfuncs}
\noindent Four loss functions, namely, reconstruction loss, perceptual loss, edge loss \cite{edgeloss}, and FFT loss\cite{fftloss}, have been used to supervise the task of image restoration.

\noindent The total loss L is defined in Equation~\ref{tloss}, where $\lambda_1 = 0.04$, $\lambda_2 = 1$ and $\lambda_3 = 0.01$.

\begin{equation}
    \label{tloss}
    L = L_s + \lambda_1 L_p + \lambda_2 L_e + \lambda_3 L_f
\end{equation}

\subsubsection{Reconstruction Loss:}

\noindent The restored clear output image is compared with its ground truth value in the spatial domain using a standard $l_1$ loss as demonstrated in Equation~\ref{L1loss}. We use $l_1$ loss instead of $l_2$ loss as it does not over-penalize the errors and leads to better image restoration performance \cite{zhao2016loss}.

\begin{equation}
    \label{L1loss}
    L_s = \frac{1}{N}\sum\limits_{i=1}^{n}\parallel x_{i}^{gt} - NAFNet(x_{i}^{dark, hazy})\parallel_{1}
\end{equation}

\noindent In the above equation, $x_{i}^{gt}$ refers to the ground truth clear image, and $NAFNet(x_{i}^{dark, hazy})$ denotes the output of our proposed NAFNet when a dark and hazy image is fed to the network.

\subsubsection{Perceptual Loss:}

\noindent To reduce the perceptual loss and improve the image's visual quality, we utilize the features of the pre-trained VGG-19 network \cite{vgg} obtained from the output of one of the ReLU activation layers. It is defined in Equation~\ref{ploss}, where $w_{ij}$, $h_{ij}$, and $c_{ij}$ refer to the dimensions of the respective feature maps inside the VGG-19 architecture. $\phi_{ij}$ denotes the feature maps outputted from the jth convolutional layer inside the i-th block in the VGG network.
\begin{equation}
    \label{ploss}
    L_p = \frac{1}{w_{ij}h_{ij}c_{ij}}\sum\limits_{x=1}^{w_{ij}}\sum\limits_{y=1}^{h_{ij}}\sum\limits_{z=1}^{c_{ij}}\parallel \phi_{ij}(I_{gt})_{xyz} - \phi_{ij}(I_{out})_{xyz} \parallel
\end{equation}

\subsubsection{Edge Loss:}
To recover the high-frequency details lost because of the inherent noise in dark and hazy images, we have an additional edge loss to constrain the high-frequency components between the ground truth and the recovered image.

\begin{equation}
\label{eloss}
    L_e = \sqrt{(\nabla^{2}(I_{gt}) - \nabla^{2}(I_{out}))^2 + \epsilon^2}
\end{equation}

\noindent In Equation~\ref{eloss}, $\nabla^2$ refers to the Laplacian operation \cite{laplacian}, which is then applied to the ground truth and the predicted clean image to get the edge loss.

\subsubsection{FFT Loss:}
To supervise the haze-free results in the frequency domain, we add another loss called  Fast Fourier transform (FFT) loss (denoted by $L_f$ in Equation~\ref{floss}. It calculates the loss of both amplitude and phase using the $l_1$ loss function without additional inference cost.

\begin{equation}
    A_{x_{i}^{gt}}, P_{x_{i}^{gt}} = FFT(x_{i}^{gt}),
\end{equation}

\begin{equation}
    A_{x_{i}^{out}}, P_{x_{i}^{out}} = FFT(x_{i}^{out}),
\end{equation}

\begin{equation}\label{floss}
    L_f = \frac{1}{N}\sum\limits_{i=1}^{n}(\parallel A_{x_{i}^{gt}} - A_{x_{i}^{out}} \parallel_{1} + \parallel P_{x_{i}^{gt}} - P_{x_{i}^{out}} \parallel_{1})   
\end{equation}

\begin{figure*}[ht]
    \centering
    \includegraphics[width = 0.9\textwidth]{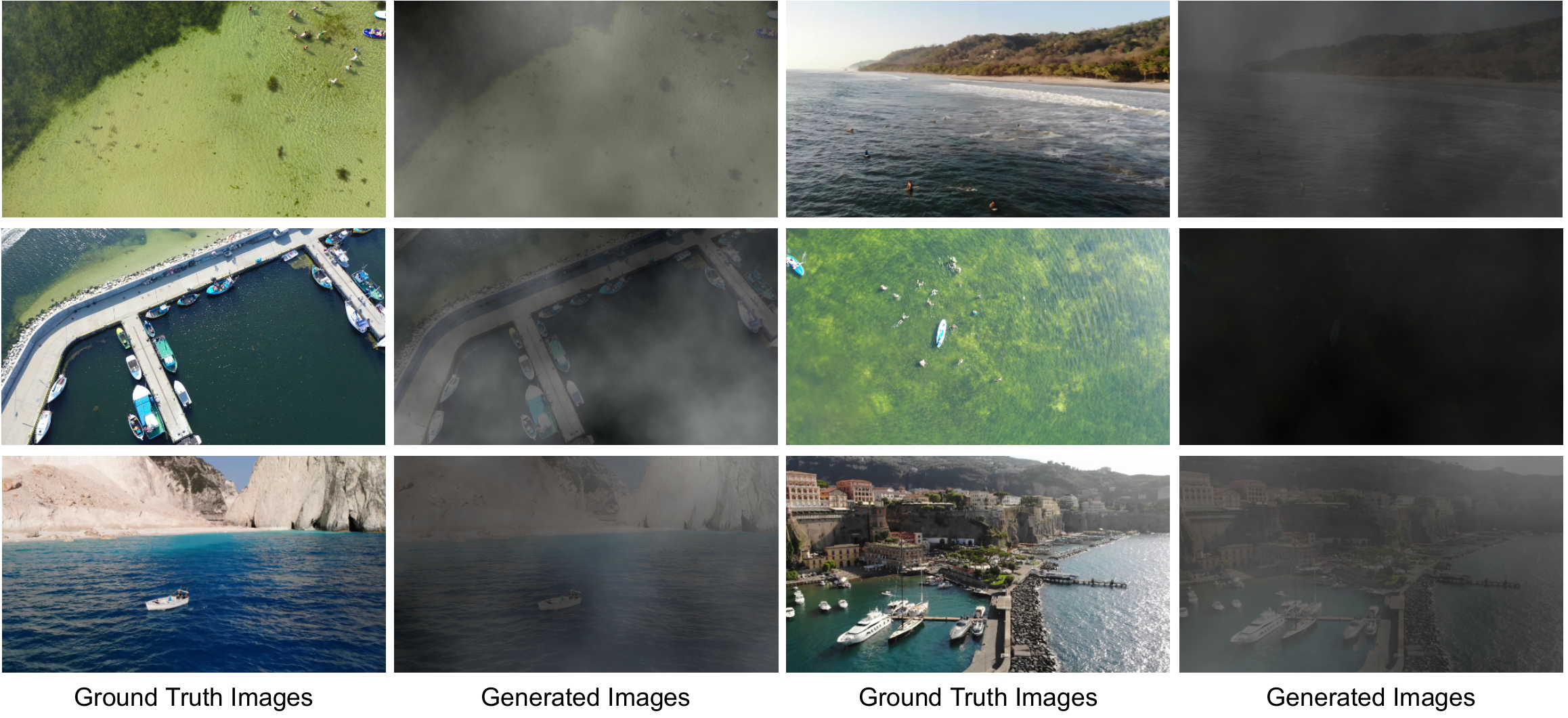}
    \caption{\textbf{Visual illustration of a few sample images from our dataset.} Columns 1 and 3 show original images taken from AFO Dataset \cite{afo}, whereas Columns 2 and 4 show their corresponding images generated as explained in Section \ref{dataset} simulating low-visibility conditions. }
    \label{fig:dataset-img}
\end{figure*}

\section{Experimental Results}
\label{eresults}

\noindent To demonstrate the outcomes of our model's approach towards image enhancement under low-visibility conditions, this section contains a detailed description of the dataset generated and used in Section~\ref{dataset}, the experimental settings in Section~\ref{settings}, the metrics used for evaluation in Section~\ref{metrics} and a discussion on the results obtained in Section~\ref{dandc} and~\ref{ablation}.

\subsection{Dataset Details}
\label{dataset}
\noindent Due to the lack of available datasets that meet our requirements, we generate a new one using the AFO dataset \cite{afo}. The dataset generation process has been elaborated below, and the final images have been shown in \figurename~\ref{fig:dataset-img}.

\begin{itemize}
    \item \textbf{Haze effect} - To  add fog, imgaug \cite{imgaug}, a well-known python library was used. A random integral value between {3, 4, 5} was selected, representing the fog's severity. For each image, this random number was chosen and pre-defined functions within the package were utilized to add a layer of fog to the image.
    
    \item \textbf{Low-light Effect} - Given a normal image, our goal is to output a low-lit image while preserving the underlying information. We follow the pipeline introduced \cite{cui2021multitask}, which parametrically models the low light-degrading transformation by observing the image signal processing (ISP) pipeline between the sensor measurement system and the final image. The low-illumination-degrading pipeline is a three-stage process:
    
    \begin{itemize}
        \item Unprocessing procedure - This part aims to synthesize RAW format images from input sRGB images by invert tone mapping, invert gamma correction, and the transformation of the image from sRGB space to cRGB space, and invert white balancing. 
        \item Low Light Corruption - This aims at adding shot and read noises to the output of the unprocessing procedure, as these are common in-camera imaging systems. Shot noise is a type of noise generated by the random arrival of photons in a camera, which is a fundamental limitation. Read noise occurs during the charge conversion of electrons into voltage in the output amplifier, which can be approximated using a Gaussian random variable with zero mean and fixed variance.
        \item ISP Pipeline - RAW image processing is done after the lowlight corruption process in the following order: add quantization noise, white balancing from cRGB to sRGB, and gamma correction, which finally outputs a degraded low-light image.
    \end{itemize}
    
    \item \textbf{Combination of Haze and Low-light Effect} - Results of implementing the low-light generation algorithm described above on foggy images generated using img-aug are shown here. It can be seen that combining the two (fog and low light) has introduced adversity in finding the location of the objects in the water bodies. Moreover, finding a unique solution for such a combination has not been explored to date

\end{itemize}

\subsection{Experimental Settings}
\label{settings}
\noindent The images were resized to get the resultant dimensions as $256 \times 456$. Adam optimizer with an initial learning rate of $1e\minus4$, $\beta_1$, and $\beta_2$ with a value of 0.9 and 0.999 were chosen. The batch size was fixed as 2. We have used 3 groups in all our experiments, each with 16 blocks. Pytorch backend was used to compile the model and train it.

\subsection{Evaluation Metrics}
\label{metrics}

\noindent We reported the results we obtained using two standard image restoration metrics (i.e., PSNR and SSIM). These metrics will help us quantitatively evaluate the performance of our model in terms of feature colors and structure similarity. High PSNR and SSIM values if indicative of good results.
\section{Experimental Results}
\label{res}
\noindent The architecture used is given in \figurename~\ref{fig:arch}. This section gives a detailed analysis of the results obtained by the proposed method.
\setlength{\tabcolsep}{0.55em}
{\renewcommand{\arraystretch}{1.5}
\small
\begin{table}[t!]
\centering
\begin{tabular}{ccccc} \toprule
    {Method} & {Year}  & {PSNR} & {SSIM} \\ \midrule
    Zero-DCE\cite{guo2020zero} & $2020$ & $12.323$ & $0.529$ \\
    SGZNet\cite{zheng2022semantic} & $2022$ & $12.578$ & $0.519$ \\
    BPPNet\cite{https://doi.org/10.48550/arxiv.2008.06713} & $2022$ & $15.507$ & $0.755$ \\
    DehazeNet\cite{cai2016dehazenet} & $2016$ & $15.710$ & $0.391$ \\
    Star-DCE\cite{zhang2021star} & $2021$ & $16.651$ & $0.539$ \\
    FFANet\cite{fftloss} & $2020$ & $15.050$ & $0.582$ \\
    MSBDN-DFF\cite{MSBDN-DFF} & $2020$ & {\color{blue}$16.686$} & {\color{blue}$0.689$}\\
    LVRNet (\textbf{Ours}) & $2022$ &  {\color{red} $25.744$}& {\color{red}$0.905$} \\
    \bottomrule
\end{tabular}
\label{table:prev-comp}
\caption{\textbf{Quantitative comparison of our proposed network with previous work.} The best results and the second-best results have been highlighted with red color and blue colors, respectively.}
\end{table}
}
\setlength{\tabcolsep}{0.55em}
{\renewcommand{\arraystretch}{1.5}
\small
\begin{table*}[t!]
\centering
\begin{tabular}{ccccccc} \toprule
    {S.no.} & {Reconstruction Loss}  & {Perceptual Loss} & {Edge Loss} & {FFT Loss} & {PSNR} & {SSIM} \\ \midrule
    1. & \cmark & \cmark & \xmark & \xmark & 24.070 & 0.870 \\
    2. & \cmark & \cmark & \xmark & \cmark & 25.455 & 0.903 \\
    3. & \cmark & \cmark & \cmark & \xmark & 25.624 & 0.897 \\
    4. & \cmark & \xmark & \cmark & \cmark & 25.719 & 0.900 \\
    5. & \cmark & \cmark & \cmark & \cmark & 25.744 & 0.905 \\
    
    \bottomrule
\end{tabular}
\label{table:abl-table}
\caption{\textbf{Ablation experiments:} We train our model using different combinations of loss functions to understand the importance of individual losses for image restoration. The best results are obtained when the model is trained using all the loss functions mentioned in this work.}
\end{table*}
}

\subsection{Discussion and Comparison}
\label{dandc}

\noindent In this subsection, we discuss the evaluation results obtained by the proposed pipeline. Previous methods were trained on the newly generated dataset and tested to compare their metrics with our model's performance. These methods were built to enhance the low-light image or obtain a clear image from a hazy one. The results are mentioned in Table 1.

\noindent We observe a huge increase in the PSNR value as compared to Zero-DCE\cite{guo2020zero}, which enhances the low-light image as a curve estimation problem. However, it introduces an even amplified noise leading to color degradation as seen in \figurename~\ref{fig:abstract}. Notwithstanding its fast processing speed, Zero-DCE has limited noise suppression and haze removal capacity. Star-DCE\cite{zhang2021star}, which uses a transformer backbone instead of a CNN one in the Zero-DCE network, shows a 35.12\% increase in PSNR value. Owing to the added LAM structure, using which our model can focus on more important feature maps, we can achieve a 54\% higher PSNR value. 

\noindent SGZNet\cite{zheng2022semantic} uses pretrained networks for enhancement factor estimation, thus their result is dependent on those pretrained weights, leading to a lower PSNR value of 12.578 on LowVis-AFO. From \figurename~\ref{fig:abstract}, we observe that the result obtained from SGZNet is still degraded by excessive noise and lacks saturation. DehazeNet\cite{cai2016dehazenet} is limited by the network's depth and cannot generalize to real-world scenarios. Hence, it results in a low PSNR of 15.710. Methods like BPPNet\cite{https://doi.org/10.48550/arxiv.2008.06713} and FFANet\cite{qin2020ffa} are end-to-end deep learning methods for image dehazing. BPPNet\cite{https://doi.org/10.48550/arxiv.2008.06713} distorts the color distribution in the recovered image as it cannot remove the dark regions, whereas FFA-Net\cite{qin2020ffa} produces image with a lower perceptual quality.

\noindent We propose an end-to-end deep learning pipeline (0.43M parameters) that can perform image dehazing and low-light image enhancement with a significant decrease in the number of parameters as compared to MSBDN-DFF \cite{MSBDN-DFF} (31M parameters) and FFA-Net\cite{qin2020ffa} (4.45M parameters). 

The supplementary material has provided a discussion on the number of parameters of other models. We also trained the model for 10 epochs with fewer NAF blocks to prove that we achieved better results than the lighter results, not due to an increase in parameters but because of the self-sufficiency of the added LAM module, non-linear activation networks, and residual connections. The results of these experiments are reported in the supplementary material.

\subsection{Ablation Studies}
\label{ablation}

\noindent To prove the importance of the perceptual loss, edge loss, and fft-loss, added to supervise the training procedure, we conducted experiments excluding each of them and reported the values of PSNR and SSIM in Table 2. We keep the $l_1$ loss function constant in all experiments as it is critical in image restoration tasks. We observe an increase in metric values in the lower rows compared to row 1. As a result of more supervision in the unchanged architecture, there is an increase in the quality of clear images obtained, which are demonstrated in the supplementary material. There is also an increase in PSNR value (which depends on per-pixel distance) in row 3, once we train the model without perceptual loss. This is seen as perceptual loss doesn't compare individual pixel values but the high-level features obtained from a pretrained network. In row 4, we get a lower PSNR value on excluding edge loss compared to row 5, as we get lesser edge supervision. Overall, we get the best performance when we include all the loss functions, as seen in row 5.


\section{Conclusion}
\noindent In this work, we have presented Low-Visibility Restoration Network (LVRNet), a new lightweight deep learning architecture for image restoration. We also introduce a new dataset, LowVis-AFO, that includes a diverse combination of synthetic darkness and haze. We also performed benchmarking experiments on our generated dataset and surpassed the results obtained using the previous image restoration network by a significant margin. Qualitative and quantitative comparison with previous work has demonstrated the effectiveness of LVRNet. We believe our work will motivate more research, focused on dealing with a combination of adverse effects such as haze, rain, snowfall, etc. rather than considering a single factor. In our future work, we plan to extend LVRNet for image restoration tasks where more factors, that negatively impact the image quality, are taken into account.
\newpage

\begin{center}
    \LARGE
    \textbf{Supplementary Material}
\end{center}
\setcounter{section}{0}

\normalsize
\noindent To make our submission self-contained and given the page limitation, this supplementary material provides additional details. 
Section~\ref{sec:graph} gives an overview of the number of parameters and PSNR obtained by different methods. Section~\ref{sec:loss} contains visual results that highlight the significance of the loss functions. Section~\ref{sec:block} contains the ablation experiment with lesser blocks, and Section~\ref{sec:lam} demonstrates the architecture diagram of the level attention module.

\section{PSNR vs Parameters}
\label{sec:graph}
\begin{figure}[ht]
    \centering
    \includegraphics[width = 0.35\textwidth]{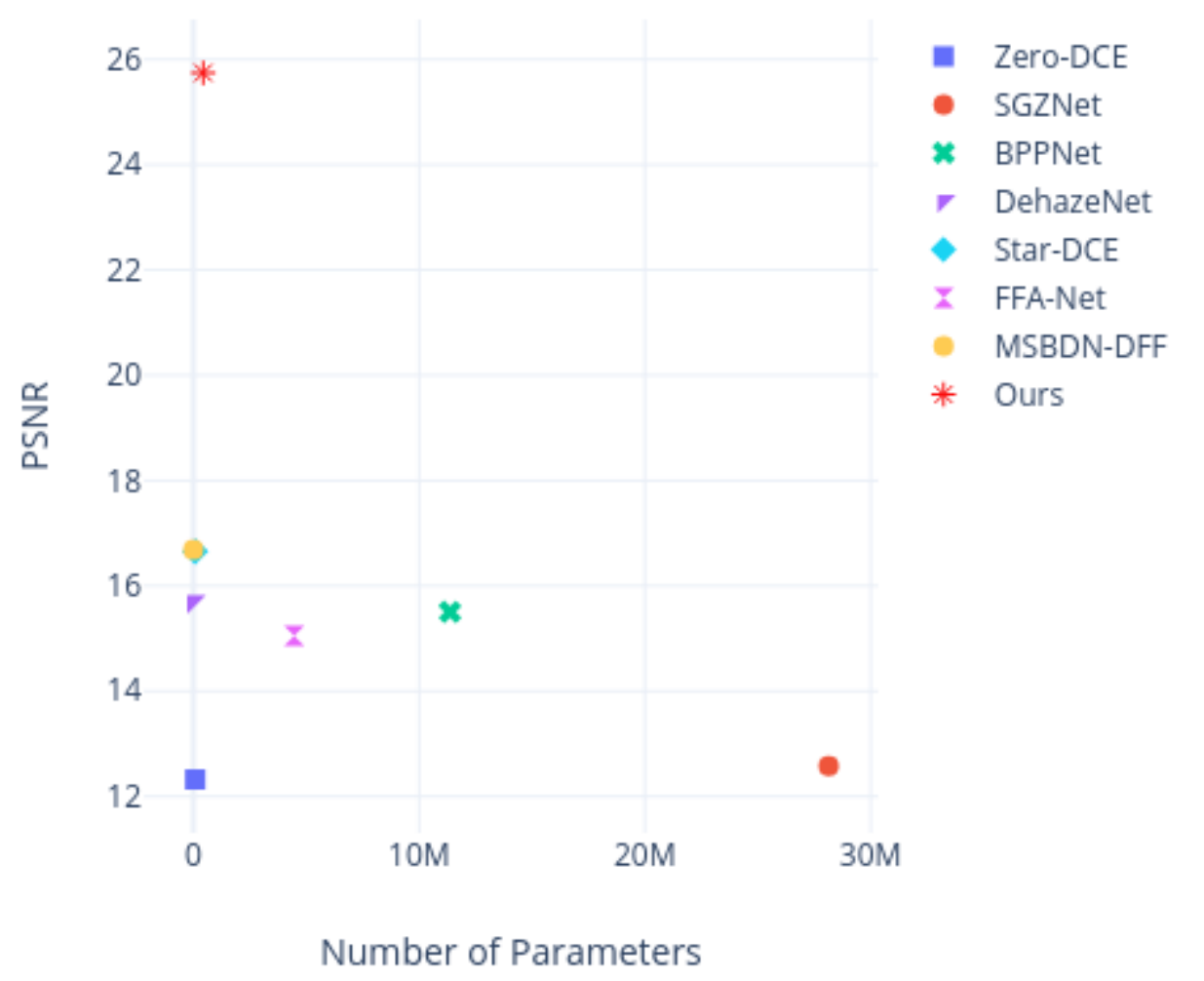}
    \caption{The PSNR vs Number of Parameters of recent image restoration methods on the newly proposed LowVis-AFO dataset.}
    \label{fig:psnr-param}
\end{figure}

\noindent \figurename~\ref{fig:psnr-param} presents the PSNR vs. Parameters plot that the previous methods and our method achieved on the testing set of LowVis-AFO. Our model outperforms the state-of-the-art image dehazing and low-light image enhancement methods by a good margin while having a lesser number of parameters.

{\renewcommand{\arraystretch}{1.5}
\begin{table}[ht]
\centering
\small
\setlength{\tabcolsep}{4 pt}
\begin{tabular}{cccccc} \toprule
     {S.no.}&{\#Blocks} & {PSNR} & {SSIM} & {\#params} & {Runtime(s)} \\ \midrule
    1. & 14 & 21.3432 & 0.8626 & 0.38M & 0.035\\
    2. & 12 & 20.4302 & 0.8488 & 0.33M & 0.029 \\
    3. & 10 & 20.2965 & 0.8494 & 0.28M & 0.024\\
    \bottomrule
\end{tabular}
\label{table:small-model}
\caption{Results of the experiments conducted on a lesser number of NAF blocks. The training was done for 10 epochs and the metrics were obtained on the test set thereafter.}
\end{table}
}
\begin{figure*}[t!]
    \centering
    \includegraphics[width=0.97\textwidth]{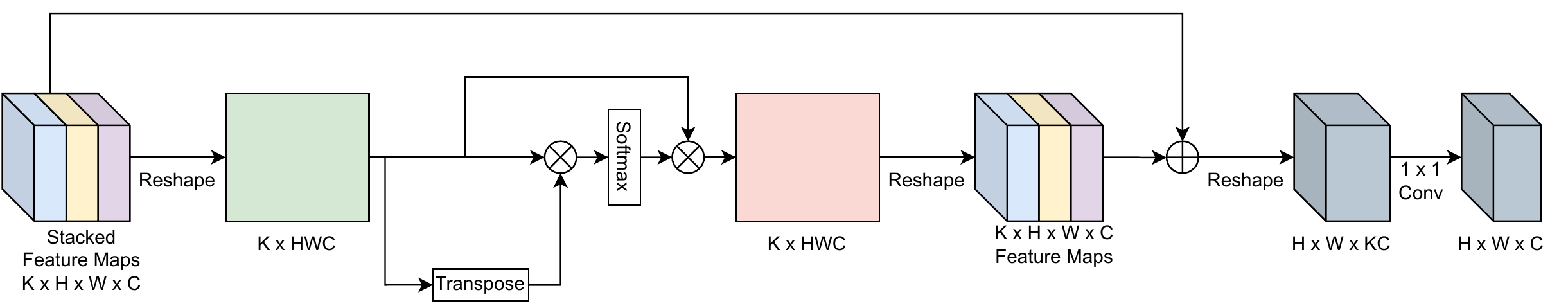}
    \caption{Visual illustration of operations performed by Level Attention Module. }
    \label{fig:lam}
\end{figure*}
\section{Ablation Experiment on Different Loss Functions}
\label{sec:loss}

\begin{figure*}[ht]
    \centering
    \includegraphics[width = \textwidth]{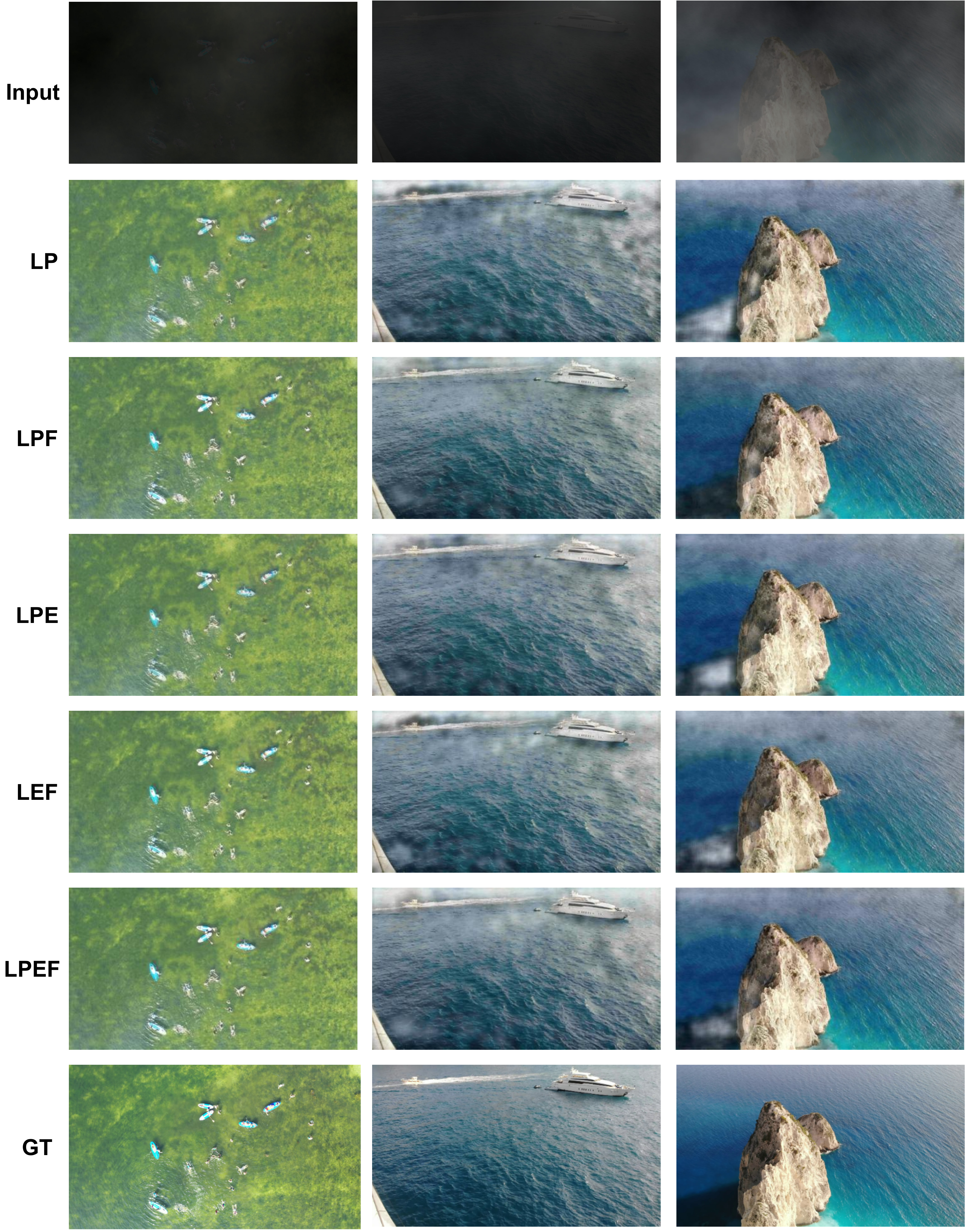}
    \caption{Qualitative results obtained from experiments conducted on different loss functions. In the figure, L = L1 loss, P = Perceptual Loss, E = Edge loss and F = FFT loss. }
    \label{fig:loss-diag}
\end{figure*}

\figurename~\ref{fig:loss-diag} demonstrates the visual results obtained when we conducted experiments excluding some loss functions. The motivation behind the experiment is to highlight the importance of the extra loss functions (perceptual loss, edge loss, fft-loss) added to supervise our pipeline. The quantitative results are given in Table 2 in the main manuscript.

\section{Ablation Experiment with Lesser Number of Blocks}
\label{sec:block}
\noindent To prove the self-sufficiency of the individual components included in our architecture such as LAM, we conduct experiments with a lesser number of NAF blocks \cite{chen2022simple} and reported the PSNR and SSIM obtained in Table 1. Seeing the results, we can conclude that our model achieves better results, not because of an increase in the number of parameters as compared to the lighter model, but because of the entire pipeline adopted.

\section{Level Attention Module}
\label{sec:lam}
\noindent As mentioned in the main text, the diagram for LAM\cite{zhang2021benchmarking} has been provided here in the supplementary material. (refer \figurename~\ref{fig:lam})

{\small
\bibliographystyle{plain}
\bibliography{main}
}
\end{document}